\documentclass[a4paper]{article}

\usepackage{INTERSPEECH2019}
\usepackage[utf8]{inputenc}

\title{Embedding-based system for the Text part of CALL v3 shared task}
\name{Volodymyr Sokhatskyi$^1$, Olga Zvyeryeva$^1$, Ievgen Karaulov$^1$, Dmytro Tkanov$^1$}
\address{
  $^1$Sciforce, Ukraine}
\email{vsokhatskyi@sciforce.solutions, ozvereva@sciforce.solutions, ekaraulov@sciforce.solutions, dtkanov@sciforce.solutions}

\begin{document}

\maketitle

\begin{abstract}

This paper presents a scoring system that has shown the top result on the text subset of CALL v3 shared task. The presented system is based on text embeddings, namely NNLM~\cite{nnlm} and BERT~\cite{Bert}. The distinguishing feature of the given approach is that it does not rely on the reference grammar file for scoring. The model is compared against approaches that use the grammar file and proves the possibility to achieve similar and even higher results without a predefined set of correct answers.

The paper describes the model itself and the data preparation process that played a crucial role in the model training.
\end{abstract}
\noindent\textbf{Index Terms}: computer assisted language learning, CALL shared task, text embeddings

\section{Introduction}

Computer Assisted Language Learning, or CALL, is ``the research for and study of applications of the computer in language teaching and learning''~\cite{Levy-1997}.

However, rapid developments in technologies and machine learning methods in  recent years have transformed CALL from a simple request-response system based on certain predefined rules to a complex Artificial Intelligence system.

In speaking practice, CALL systems utilizing the automatic speech recognition (ASR)  technology offer new abilities to process learners' responses for error detection and automated feedback generation~\cite{SLaTE-2013, ELRA-2014, LT-2014}. As an initiative to further develop the related technologies, a shared task for the spoken CALL research was presented in 2016 and  participating systems were reported in the ISCA SLaTE 2017 workshop~\cite{SLATE-2017}. The task is to provide feedback to prompt-based spoken responses by learners of English who use the CALL-SLT system~\cite{LT-2014}. Participating systems are expected to accept responses with correct meaning and language usage, and reject others. Following the success of the first and second shared tasks, the third edition with the same training data was announced in Autumn 2018 and the test data was released on April 21, 2019~\cite{Interspeech-2018}. Similarly to the previous edition, the task organizers  provide  audio  data,  ASR  outputs, and reference response grammar.

The shared task is composed of two subtasks: the text task that has the ASR outputs for the spoken responses  provided by the organizers, and the speech task in which participants can use their own recognizers to process audio responses. The performance is evaluated with the $D_{full}$ metric~\cite{Interspeech-2018}. This metric rests on three intuitions:
\begin{itemize}
	\item The system should reject incorrect answers as often as possible, and reject correct answers as seldom as possible.
	\item The more pronounced the difference between the system's response to incorrect answers and correct ones, the more useful it is.
	\item Some system errors are more critical than others. For instance, it is worse for the system to accept a sentence which is incorrect in terms of meaning than to accept the one which is correct in terms of meaning but incorrect in terms of language/grammar.
\end{itemize}

In order to prevent ``gaming'' of the metric, entries are also required to reject at least 50\% of all incorrect responses and accept at least 50\% of all correct responses. 

This paper describes the system developed by the authors for the text task. Among text task competitors it is the only submission that beats the grammar file-based baseline model. The model also surpasses the baseline even without using the grammar file, resting solely on careful data preparation.

\begin{table}[!ht]
\begin{tabular}{l|l|llllll}
Sumbission      & Task     & D      & $D_{full}$  \\
BaselinePerfRec & Text   & 10.08  & 12.327 \\
GGG             & Speech & 11.348 & 6.342  \\
HHH             & Speech & 12.75  & 6.229  \\
III             & Speech & 12.416 & 6.13   \\
OOO             & Speech & 9.401  & 5.608  \\
PPP             & Speech & 9.401  & 5.608  \\
NNN             & Speech & 8.95   & 5.476  \\
CCC             & Speech & 10.082 & \textbf{5.43}  \\
AAA             & Speech & 9.046  & \textbf{5.149}  \\
BBB             & Speech & 7.567  & \textbf{4.909}  \\
FFF             & Text   & 5.998  & \textbf{4.413}  \\
DDD             & Text   & 6.28   & \textbf{4.403}  \\
EEE             & Text   & 5.449  & \textbf{4.227}  \\
Baseline        & Text   & 5.176  & 4.09  \\
MMM             & Text   & 4.953  & 3.999  \\
KKK             & Text   & 4.822  & 3.936  \\
LLL             & Text   & 4.697  & 3.876  \\
JJJ             & Text   & 2.356  & 1.665  
\end{tabular}
\caption{Results for anonymised submissions (scores of our systems are highlighted).}
\label{tabl-fin-res}
\end{table} 

\section{Previous work}

All proposed solutions for CALL v1 and v2 relied heavily on the reference grammar file~\cite{SLaTE-2017-1, Interspeech-2018-1, Interspeech-2018-2, Interspeech-2018-3, Interspeech-2018-4, Interspeech-2018-5, Interspeech-2018-6}. For example, one of the last year's submissions~\cite{Interspeech-2018-3} processed the ASR output and up to 10 entries from the reference grammar file using the \textbf{doc2vec} model. Afterwards, they used the word mover distance to get 10 distances representing the student's utterance. The distances were passed as inputs to a neural network that generated the final decision score.

In our opinion, using reference grammar files makes the scoring system inherently non-scaleable. It becomes hard to transfer results to other language pairs or even to other prompts for the same pair, as it automatically implies extending the reference grammar file with each new prompt. Our goal was to create a scoring system that would not rely on reference grammar files and would achieve comparable results as solutions based on grammar files.

\section{Dataset}

\subsection{Overview}

The data provided for the third edition of CALL shared task was collected from an online CALL tool used to help young Swiss German students improve their English fluency. The training data was the same as the data provided for the second edition of the task. Each participant was asked to respond verbally in English to a given German text prompt. Then each response is labeled as ``correct'' or ``incorrect'' for its linguistic correctness (language) and its meaning. A response is accepted when it is correct in both language and meaning given the prompt. Otherwise, it is rejected. It is possible that a response is correct only in one aspect. The following  shows an example of prompt in German and accepted student's response: 

{\bf Prompt} \textit{Frag: Wo ist Piccadilly Circus?}

{\bf Response} \textit{Where is Piccadilly Circus?}

For the best quality, each utterance was processed with four of the best assessment systems from the first shared task~\cite{SLaTE-2017-1, SLaTE-2017-2, SLaTE-2017-3, SLaTE-2017-4} to obtain accept/reject decisions for the language criterion. According to the results and to the judgement provided by three native English speaker experts familiar with the domain, the dataset of 6698 utterances was divided into three groups: A, B and C of descending reliability (with A the highest and C the lowest).

Notably, the recording environment is not perfect due to the background noises in schools. It affects ASR transcripts and results in noise in labels. Mostly it concerns Group C, but Groups A and B have a number of noisy utterances as well. 

Group C is dataset with ambiguous judgement mostly due to bad quality of recordings. In our experiments we observed, that inclusion of group C usually resulted in decrease of accuracy of final scoring system.

A new test dataset containing 1000 utterances was released in April 21, 2019. 

\subsection{Challenges}

While working on the scoring system, we encountered several issues in the dataset. According to the CALL shared task rules for the text scoring system, we used ASR transcripts provided by the organazires. Therefore, the following challenges applied to it.  

First, the dataset contains few entries that are quite noisy. We also used data from the first edition of the Spoken CALL Shared Task, but it often had ambiguous judgments. In other words, entries in the text task might have different labels for the same or similar ASR transcript.

To illustrate the point, the following entries in the first dataset were labeled sometimes as ``correct'' and sometimes as ``incorrect'' by to language criteria: 

{\bf Prompt} \textit{Frag: Ich möchte die Dessertkarte}

{\bf Transcription} \textit{I want the desert menu}

{\bf RecResult} \textit{I want the dessert menu}

Second, in the training set, different audio recordings of the same phrase may have the same ASR transcript. It might be a benefit for the speech task, yet for the text task it resulted in a lot of duplicated entries. After removing all of them from the training data, we had a dataset of only about 2000 utterances. Further elimination of Group C of the train set resulted in even smaller dataset.

For example, Group A of training data has 22 duplicate entries of:

{\bf Prompt} \textit{Frag: Gibt es ein Hotelrestaurant?}
 
{\bf Transcription} \textit{Is there a hotel restaurant}

and 22 duplicate entries of:

{\bf Prompt} \textit{Sag: Ich habe keine Reservation}
 
{\bf Transcription} \textit{I have no reservation}

Third, there are many duplicates of utterances in both 2nd and 3rd edition test sets for the text task. However, the corresponding audio recordings are unique, which makes such entries useful for the speech task, but complicates the text task. Furthermore, these test sets intersect with the training set. This leaves only about 300 unique entries out of 1000. So, there is a danger to create a system with seemingly acceptable performance that would merely overfit the training dataset.

It might be reasonable to keep separate datasets for the text and speech subtasks. Otherwise, the textual part requires significant preprocessing to eliminate duplicates within and between test and train sets.

\section{Text scoring system}
\subsection{Dataset resampling}
One of our key efforts was to form a high-quality training set.

First, we improved the reference grammar file by removing a number of entries with mistakes. For example, there are 15 {\it ``can I pay with credit card''} entries in test set for the second CALL shared task, as well as many similar ones like {\it ``I would like to pay with credit card''}. The correct phrase, according to the training set, is {\it 'I would like to pay with \textbf{a} credit card'} -- missing article 'a' should result in the ``incorrect'' judgement.

Then we concatenated datasets A and B and the part of CALL 1st edition training dataset that was correctly judged by the baseline grammar.

Also, we merged the columns {\it RecResult} and {\it Transcription}. Therefore, each entry from the data set was divided into two entries: {\it RecResult} and {\it Transcription} with the same judgements.

Then the following preprocessing steps were taken. They are listed here for completeness and generally follow the approach of~\cite{Interspeech-2018-1} and ~\cite{Interspeech-2018-2}.
\begin{itemize}
	\item All irregular white-space is removed and replaced with a single empty space.
	\item The artifacts of the ASR system (``ah'', ``um'', ``euhm'', ``ggg'' etc.) are removed.
	\item Superfluous  words  like ``yes'', ``thanks'', ``thank you'', ``please'' and ``also'' are removed as they have no influence on the meaning and linguistic correctness, except the cases where they are the only word in the entry.
	\item Words like ``no'' or ``and'' are removed, if they are at the beginning of the sentence. Additionally, words at the end of sentences like ``no'' and ``is'' that provide neither syntax nor semantic content, are removed, as they are usually artifacts of the ASR system for noisy input.
	\item Word and phrase duplications due to false starts or repetitions are removed during the preprocessing phase.
	\item Verbs' contraction forms are replaced with their complete-word forms. For example, ``I'd'', ``they're'' and ``wanna'' are replaced with ``I would'', ``they are'' and ``want to'' respectively. 
\end{itemize}

As the last step, duplicates were removed. The final training dataset consisted of 4481 entries. In our experiments, this dataset configuration yields more consistent and higher results, than the original training set.

\subsection{Word and phrase embeddings}

The cornerstone component of our scoring system is the text embeddings estimator. Previous works also relied on embeddings in form of doc2vec. Our main contribution is that we used only embeddings information at the inference stage.

We used Bidirectional Encoder Representations from Transformers -- {\bf BERT}~\cite{Bert}, more specifically, the {\it multi\_cased\_L-12\_H-768\_A-12} model trained on Wikipedia and the BookCorpus. We did not finetune BERT, because, in our opinion, the amount of data is too limited for that. Though, with proper data augmentation, it might be reasonable to try finetuning.

In addition to {\bf BERT}~\cite{Bert}, we tried to use other models for embeddings generation, namely {\bf nnlm}~\cite{nnlm}, {\bf elmo}~\cite{elmo}, {\bf doc2vec}~\cite{doc2vec}, {\bf word2vec}~\cite{word2vec} and {\bf universal-sentence-encoder}~\cite{use}. Among tested alternatives, {\bf nnlm} appeared to be superior to other models in the context of CALL shared task. The {\bf nnlm} is a neural network-based language model~\cite{nnlm}. It allows mapping words to 50-dimensional embedding vectors. When it was necessary, we aggregated sequences of word vectors into phrase vectors.

The first approach we tried with word embeddings was the approach similar to previous year's winner~\cite{Interspeech-2018-2}. We calculated the similarity between students' responses and corresponding entries from the reference grammar. We ran several experiments with different ways to measure similarity: cosine similarity between phrase vectors, DTW distance, word mover distance, etc. Every experiment resulted in scores lower than the baseline grammar system.

The best results were achieved using {\bf BERT} and {\bf nnlm}. {\bf BERT} produces contextual embeddings, so we expected high performance. In the {\bf nnlm}-basesystem, word embeddings are averaged into a sentence embedding, so it does not take the word order into account. In this context, the relatively high performance of this model is surprising.

From {\bf BERT}, we obtained a 768-dimensional vector for each phrase from the dataset. We used German prompts translated using the Google Translate service and corresponding English answers concatenated via $'|||'$ as inputs. In sentence pair tasks typical for {\bf BERT} model, such as question answering and entailment, Sentence A is separated from Sentence B with the $'|||'$ delimiter. This approach turned out to work well in our case. From {\bf nnlm} we obtained a 50-dimensional vector per input phrase. We used two models for {\bf nnlm}: for original German prompts we used a German model trained on the German Google News 30B corpus, while for responses and for English translations of prompts we used an English model trained on the English Google News 7B corpus. The model with the highest capacity among those we trained used 918-dimensional inputs: 768 from \textbf{BERT} and 3 x 50 from \textbf{nnlm}.

\subsection{Training}

We trained a model to solve the classification problem. Each sample belonged to one of three mutually exclusive classes: correct, wrong language, wrong meaning.
As input for the model, we used 918-dimensional vectors that contained a 768-dimensional embedded vector from \textbf{BERT} and a 150-dimensional vector from \textbf{nnlm} based on German prompts, German translated prompts and students' responses. We trained a neural network with a single hidden layer of 128 neurons with ReLU activation. For regularization, we used dropout and early stopping. Our training loss (cross-entropy) showed a different behavior than target metrics ($D_{full}$). One of the most important points, therefore, was to run early stopping over $D_{full}$. Finally, to get more robust results, we used an ensemble of models -- averaged outputs of models trained on different parts of the trainset and initialized with differnet random states.

\subsection{Results}

To get estimates of final scores of our submissions, we performed validation on the test set from the 2nd Edition Spoken CALL SharedTask. The results are presented in Table~\ref{tabl-res}:

\begin{table}[!ht]
\begin{tabular}{l|lllll}
Model        & Pr    & Rec    & F     & D    & $D_{full}$ \\
BERT         & 0.958 & 0.876 & 0.915 & 6.70 & 5.89    \\
BERT+        & 0.958 & 0.885 & 0.920 & 7.24 & 6.16    \\
nnlm         & 0.965 & 0.875 & 0.917 & 6.88 & 6.62    \\
nnlm+        & 0.964 & 0.885 & \textbf{0.923} & \textbf{7.43} & \textbf{6.67}    \\
BERT + nnlm  & 0.958 & 0.879 & 0.917 & 6.85 & 5.96    \\
BERT + nnlm+ & 0.958 & \textbf{0.887} & 0.921 & 7.33 & 6.20    \\
Grammar  & 0.936 & 0.872 & 0.903 & 6.07 & 4.87    \\
Updated grammar  & \textbf{0.966} & 0.872 & 0.917 & 6.72 & 6.46    
\end{tabular}
\caption{Comparison of text scoring models.
{\bf BERT} -- model, based only on BERT-embedding vectors.
{\bf BERT+} -- model, based on BERT-embedding vectors and Updated grammar: if Updated Grammar judges an entry as correct, we accept the answer, otherwise we use model for judgment. 
{\bf nnlm} -- model, based only on nnlm-embedding vectors.
{\bf nnlm+} -- model, based on nnlm-embedding vectors and Updated grammar.
{\bf BERT + nnlm} -- model, based on BERT-embedding vectors concatenated with nnlm-embedding vectors.
{\bf BERT + nnlm+} -- model, based on the previous model and Updated grammar.
Updated grammar is the grammar file provided by the organizers without several entries that contained mistakes. For more details, see Dataset resampling subsection.}
\label{tabl-res}
\end{table}

The results suggest that the {\bf nnlm+} model is superior to others, though the difference between {\bf nnlm} and {\bf nnlm+} is subtle. Only two models outperform the updated grammar system. After experimenting with other types of architectures, we concluded that the use of grammar file usually yields results very similar to the baseline system. In other words, almost any approach would get satisfactory results, yet it would be very hard to reach anything beyond the baseline.

On the CALL v3 test set~(Table~\ref{tabl-fin-res}), the model {\bf nnlm+}~(FFF) achieves the best performance by the $D_{full}$ metric as well. However, the model {\bf BERT + nnlm}~(DDD) shows better score than {\bf nnlm}~(EEE).

\section{Discussion}
In our opinion, the allowance of grammar file renders text subtask unattractive in comparison to audio subtask. The reason is that any increase in ASR performance would result in much more noticeable score improvements. The grammar file provides a ``low hanging fruit'' that gives results that are hard to improve upon. As a result, the work on text part of the shared task becomes implicitly penalized. We'd suggest for the next year's competition to form separate datasets for audio and text subtasks and either not provide a grammar file or to form a test set from entries that are mostly absent in the grammar file.

The work presented in this paper would be orthogonal to improvements in the ASR system. Thus, combining the described text scoring approach with one of the top performing ASRs from the speech task may yield better results than any of the two systems separately.

\section{Conclusions}

In this paper we presented a text-based scoring system for CALL v3 shared task. We also discussed the dataset and proposed changes to data formation routines for future competitions.

Our best submission to the challenge obtained $D_{full}$ score of $4.192$. The system achieved such result using {\bf nnlm} and the updated grammar. Two other submissions, {\bf BERT + nnlm} with the $D_{full}$ score of $4.178$ and {\bf nnlm} with the score of $4.025$, showed slightly worse results, but still better than the grammar baseline.

In our opinion, in spite of the slightly worse results, the last two submissions are more valuable because the corresponding systems achieved high scores without using grammar file. Hence, these systems can be easily extended to other domains and languages.

\section{Acknowledgments}
We would like to thank Andrey Osetrov for his valuable comments and suggestions.

\bibliographystyle{IEEEtran}

\begin{thebibliography}{9}

 \bibitem[1]{nnlm}
     Y.\ Bengio, R.\ Ducharme, P.\ Vincent, C.\ Jauvin, 
    '' A Neural Probabilistic Language Model,'' in textit{Journal of Machine Learning Research,} 3:1137-1155, 2003.
 \bibitem[2]{Bert}
    J.\ Devlin, M.\ W.\ Chang, K.\ Lee K.\ Toutanova, 
    ''BERT: Pre-training of Deep Bidirectional Transformers for Language Understanding,'' in textit{Google AI Language,} 2018.

 \bibitem[3]{Levy-1997}
   M.\ Levy,
   ``Computer-assisted language learning: Context and con-ceptualization.,''
   in \textit{Oxford University Press.}, 1997.
   
 \bibitem[4]{SLaTE-2013}
 	B.\ Penning de Vries, S.\ Bodnar, C.\ Cucchiarini, H.\ Strik, and R.\ v.Hout,
   ``Spoken  grammar  practice  in  an  ASR-based  CALL  sys-tem,''
   in \textit{Speech and Language Technology in Education (SLaTE), Grenoble, France, pp. 60–65,} 2013.

 \bibitem[5]{ELRA-2014}
 C.\ Cucchiarini, S.\ Bodnar, B.\ Penning de Vries, R.\ V.\ Hout, and H.\ Strik,
   ``ASR-based CALL Systems and Learner Speech Data: New Resources and Opportunities for Research and Developmentin Second Language Learning,''
   in \textit{EuropeanLanguage Resources Association (ELRA), Reykjavik, Iceland,} May 2014. [Online]. Available: https://archive-ouverte.unige.ch/unige:42119

 \bibitem[6]{LT-2014}
   E.\  Rayner,  N.\  Tsourakis,  C.\  Baur,  P.\  Bouillon,  and  J.\  Gerlach,
   ``CALL-SLT:  A  Spoken  CALL  System  Based  on  Grammarand   Speech   Recognition,'' Linguistic Issues in
   \textit{LanguageTechnology, vol. 10, no. 2,} 2014.
	
 \bibitem[7]{SLATE-2017}
 	C.\ Baur,  C.\ Chua,  J.\ Gerlach,  M.\ Rayner,  M.\ Russell,  H.\ Strik, X.\  Wei, 
   ``Overview  of  the  2017  Spoken  CALL  SharedTask,''
   in \textit{Proc.  7th  ISCA  Workshop  on  Speech  and  LanguageTechnology in Education, pp. 71–78}, 2017. [Online]. Available:http://dx.doi.org/10.21437/SLaTE.2017-13.
 
  \bibitem[8]{Interspeech-2018}
 C.\ Baur, A.\ Caines, C.\ Chua, J.\ Gerlach, M.\ Qian, M.\ Rayner, M.\ Russell, H.\  Strik and X.\ Wei,
   ``Overview of the 2018 Spoken CALL Shared Task,''
   in \textit{Interspeech 2018}, India Sep. 2018.

 \bibitem[9]{SLaTE-2017-1}
   M.\ Qian, X.\ Wei, P.\ Jancovic M.\ Russell,
   ``The University of Birmingham 2017 SLaTE CALL Shared Task Systems,''
   in \textit{Proceedings of the Seventh SLaTE Workshop,} Stockholm,  Sweden 2017.

 \bibitem[10]{SLaTE-2017-2}
   A.\ Magooda and D.\ Litman, 
	''Syntactic and semantic features forhuman like judgement in spoken call,''   
   in \textit{Proceedings of the Seventh SLaTE Workshop,} Stockholm, Sweden, 2017.

 \bibitem[11]{SLaTE-2017-3}
    Y.\ R.\ Oh, H.-B.\ Jeon, H.\ J.\ Song, B.\ O.\ Kang, Y.-K.\ Lee, J.-G.\ Park, Y.-K.\ Lee, 
    ''Deep-Learning based automatic spon-taneous speech assessment in a data-driven approach for the 2017SLaTE CALL Shared Challenge,'' in textit{Proceedings of the Seventh SLaTE Workshop,} Stockholm, Sweden, 2017.
    
  \bibitem[12]{SLaTE-2017-4}
    K.\ Evanini, M.\ Mulholland, E.\ Tsuprun, and Y.\ Qian, 
    ''Using anautomated content scoring system for spoken CALL responses: The ETS submission for the Spoken CALL Challenge,'' in textit{Proceedings of the Seventh SLaTE Workshop,} Stockholm, Sweden, 2017. 
    
    \bibitem[13]{Interspeech-2018-1}
    D.\ Julg, M.\ Kunstek, C.\ Freimoser, K.\ Berkling, M.\ Qian, 
    ''The CSU-K Rule-Based System for the 2nd Edition Spoken CALL SharedTask,'' in textit{Interspeech 2018,} Hyderabad, India, 2018.
   
   \bibitem[14]{Interspeech-2018-2}
    H.\ Nguyen, L.\ Chen, R.\ Prieto, C.\ Wang, and Y.\ Liu, 
    ''Liulishuo's System for the Spoken CALL Shared Task 2018,'' in textit{Interspeech 2018,} Hyderabad, India, 2018.
    
    \bibitem[15]{Interspeech-2018-3}
    C.\ Freimoser, M.\ Kunstek, D.\ Jülg, K.\ Berkling, M.\ Qian, 
    ''The CSU-K DNN-Based System for the 2nd Edition Spoken CALL Shared
Task,'' 2018.
    
	\bibitem[16]{Interspeech-2018-4}
    M.\ Ateeq, A.\ Hanani, A.\ Qaroush, 
    ''An Optimization Based Approach for Solving Spoken CALL Shared Task,'' in textit{Interspeech 2018,} Hyderabad, India, 2018.    
    
    \bibitem[17]{Interspeech-2018-5}
    M.\ Qian, X.\ Wei, P.\ Jancovic, M.\ Russell, 
    ''The University of Birmingham 2018 Spoken CALL Shared Task Systems,'' in textit{Interspeech 2018,} Hyderabad, India, 2018.
    
    \bibitem[18]{Interspeech-2018-6}
    K.\ Evanini, M.\ Mulholland, R.\ Ubale, Y.\ Qian, R.\ Pugh, V.\ Ramanarayanan, A.\ Cahill, 
    ''Improvements to an Automated Content Scoring System for Spoken CALLResponses: The ETS Submission to the Second Spoken CALL Shared Task,'' in textit{Interspeech 2018,} Hyderabad, India, 2018.
    
    \bibitem[19]{elmo}
    M.\ E.\ Peters, M.\ Neumann, M.\ Iyyer, M.\ Gardner, C.\ Clark, K.\ Lee, L.\ Zettlemoyer, 
    ''Deep contextualized word representations,'' arXiv:1802.05365, 2018.
    
    \bibitem[20]{doc2vec}
    Q.\ Le, T.\ Mikolov 
    ''Distributed Representations of Sentences and Documents,'' arXiv:1405.4053v2 22 May 2014.
    \bibitem[21]{word2vec}
    T.\ Mikolov, K.\ Chen, G.\ Corrado, J.\ Dean, 
    ''Efficient Estimation of Word Representations in Vector Space,'' arXiv:1301.3781v3 7 Sep 2013.
    
    \bibitem[22]{use}
    D.\ Cer, Y.\ Yang, S.\ -yi \ Kong, N.\ Hua, N.\ Limtiaco, R.\ St.\ John, N.\ Constant, M.\ Guajardo-Céspedes, S.\ Yuan, C.\ Tar, Y.-H.\ Sung, B.\ Strope, R.\ Kurzweil, 
    ''Universal Sentence Encoder,'' arXiv:1803.11175, 2018.
        
 \end{thebibliography}

\end{document}